\documentclass[letterpaper]{article} 
\usepackage{aaai2027}  
\nocopyright
\usepackage[hyphens]{url}  
\usepackage{graphicx} 
\usepackage{natbib}  
\usepackage{caption} 
\usepackage{algorithm}
\usepackage{algorithmic}
\usepackage{subfigure}

\usepackage{newfloat}
\usepackage{listings}
\DeclareCaptionStyle{ruled}{labelfont=normalfont,labelsep=colon,strut=off} 
\floatstyle{ruled}
\newfloat{listing}{tb}{lst}{}
\floatname{listing}{Listing}

\usepackage{booktabs}

\usepackage{amsmath,amssymb,bm}
\usepackage{multirow}

\title{RAMamba-Net: A Reliability-Aware and Mamba-Based Multimodal Fusion Network for Auditory Attention Detection}
\author{
    Xingyi He\textsuperscript{\rm 1}\equalcontrib,
    Ziwei Wang\textsuperscript{\rm 1}\equalcontrib,
    Dongrui Wu\textsuperscript{\rm 1}
}
\affiliations{
    \textsuperscript{\rm 1}School of Artificial Intelligence and Automation, Huazhong University of Science and Technology, China\\
    \{xyhe, vivi, drwu\}@hust.edu.cn
}

\begin{document}

\maketitle

\begin{abstract}
Auditory attention decoding (AAD) identifies the attended speaker from physiological signals, supporting neuro-steered hearing devices and natural human-machine interaction. Electroencephalography (EEG) is the dominant modality for AAD but provides incomplete evidence in naturalistic audio-visual scenes, motivating EEG and electrooculography (EOG) fusion. Existing approaches remain limited by weak cross-modal interaction, inefficient temporal modeling, and low robustness to sample variations. To address the limitations, we propose RAMamba-Net, a reliability-aware Mamba-based multimodal fusion network for AAD. RAMamba-Net employs a Mamba-enhanced band-aware convolutional Transformer to capture band-specific EEG patterns and long-range temporal dynamics. A dual-branch temporal-spatial encoder models EOG temporal and inter-channel dependencies. Cross-modal attention enables explicit modality interaction. Then, a reliability-aware module is introduced to estimate sample-wise modality weights for feature and prediction consistency, thereby enhancing multimodal fusion. Experiments on two AAD benchmarks demonstrate that RAMamba-Net effectively exploits complementary EEG-EOG information, yielding accuracy gains of 5.76\% over unimodal baselines, together with more robust decoding and discriminative representations. Further analyses show that explicit cross-modal interaction improves multimodal alignment, while the reliability-aware module suppresses unreliable modality evidence and is robust to signal perturbation and parameter variation.

\end{abstract}


\section{Introduction}
Understanding a target speaker in a crowded acoustic scene remains a major challenge for people with hearing impairment \cite{wilroth2025improving}, commonly known as the cocktail-party problem \cite{cherry1953some}. Auditory attention decoding (AAD) infers the attended speaker from physiological signals \cite{mesgarani2012selective,ding2012emergence,osullivan2015attentional}, supporting neuro-steered hearing devices, target-speaker enhancement, and natural human-machine interaction \cite{alickovic2019tutorial,geirnaert2021eeg}. These applications require reliable decoding under realistic and non-ideal recording conditions.

Electroencephalography (EEG) is the dominant AAD modality. It reflects cortical activity associated with attentional selection. EEG decoders have progressed from convolutional direction localization \cite{vandecappelle2021eeg} to spatio-temporal attention modeling \cite{su2022stanet} and end-to-end cross-subject decoding \cite{nguyen2025aadnet}. However, EEG has a low signal-to-noise ratio, is highly susceptible to artifacts \cite{somers2018generic}, and often generalizes poorly across subjects and sessions \cite{puffay2023relating}. EEG alone may therefore provide incomplete evidence of attentional behavior in naturalistic audio-visual scenes, motivating the integration of complementary physiological signals \cite{wang2025cst}.

Eye movements provide behavioral information related to auditory attention, while electrooculography (EOG) directly records ocular activity. Prior work has shown that gaze follows attended natural speech \cite{gehmacher2024eye} and that visual speech cues enhance neural tracking of the attended speaker \cite{fu2019congruent}. These findings suggest that EOG can complement EEG. Nevertheless, ocular activity may introduce confounding patterns and bias spatial-attention decoding \cite{rotaru2024what}. EOG should therefore be explicitly modeled and adaptively integrated with EEG rather than discarded as noise or combined through na\"ive concatenation.

EEG-EOG fusion for AAD still faces key challenges. First, continuous cross-modal dependencies remain insufficiently modeled. Cross-attention has proven effective in related physiological and multimodal tasks \cite{lu2015combining,yin2026multimodal,zhuang2026catnet,tsai2019multimodal}, yet bidirectional token-wise interaction between EEG and EOG remains underexplored in AAD. Second, efficient temporal modeling remains difficult. Convolutions emphasize local patterns, whereas self-attention incurs high computational cost for long sequences. Mamba provides selective state-space modeling with linear-time complexity \cite{gu2023mamba} and has been introduced into AAD models \cite{zhang2024swim,fan2025seeing}. However, the band-aware EEG encoder and cross-modal interaction module remain limited. Third, modality quality varies across samples because of artifacts, subject state, and ocular dynamics. Fixed weighting and direct concatenation cannot adapt to these variations, motivating sample-wise reliability estimation.

To address these challenges, we propose RAMamba-Net, a reliability-aware Mamba-based multimodal fusion network for AAD. RAMamba-Net integrates modality-specific encoding, cross-modal interaction, and sample-wise reliability estimation. Our contributions are summarized as follows:
\begin{itemize}
    \item We propose a unified EEG-EOG multimodal framework for AAD that integrates modality-specific encoding, cross-modal interaction, and reliability-aware fusion into an end-to-end decoding pipeline.
    \item We design a Mamba-enhanced band-aware EEG encoder that couples band-specific modeling with selective state-space learning to capture spectral patterns and long-range temporal dynamics.
    \item We introduce cross-modal attention to enable token-wise information exchange between EEG and EOG while preserving their modality-specific characteristics.
    \item We develop a reliability-aware module (RAM) to estimate sample-wise modality weights from modality preservation and cross-modal consistency in the representation and prediction spaces. Extensive experiments on two AAD benchmarks demonstrate improved decoding accuracy and representation quality, together with robust performance under signal perturbation and parameter variation.
\end{itemize}

\section{Related Work}
\subsection{EEG-Based Auditory Attention Decoding}
EEG-based auditory attention decoding has progressed from handcrafted pipelines to end-to-end models that jointly capture spectral, spatial, and temporal neural patterns. MBSSFCC \cite{jiang2022mbssfcc} combines multi-band differential-entropy topographies with ConvLSTM for spatio-temporal attention representation. BSAnet \cite{cai2023bsanet} adopts a biologically inspired spiking attention network to model attention-related EEG activity. DBPNet \cite{ni2024dbpnet} employs parallel temporal-attentive and frequency-residual branches to capture dynamic and multi-band spectral-spatial information. DARNet \cite{yan2024darnet} integrates spatio-temporal construction with dual-attention refinement for spatial distribution and long-range dependency modeling. DHGCN \cite{zhou2025dhgcn} constructs temporal and spatial hypergraphs to learn higher-order dependencies across time points and channels. FAConformer \cite{wang2026faconformer} combines band-specific CNN-Transformer encoders, frequency-aware attention, and band-wise supervision to learn within-band representations and adaptive cross-band interactions.

Despite these advances, EEG remains susceptible to low signal-to-noise ratios and environmental artifacts, which hinder stable representation learning under varying recording conditions. Moreover, EEG-only decoding may not fully characterize auditory attention in realistic audio-visual scenes. These limitations motivate multimodal AAD models to exploit complementary information while preserving robust neural representations.

\subsection{Multimodal Auditory Attention Decoding}
EOG signals provide behavioral evidence complementary to neural activity and have gained increasing attention in multimodal AAD. Gehmacher \textit{et al.} \cite{gehmacher2024eye} combined eye tracking with magnetoencephalography and showed that gaze followed attended natural speech, and was associated with attention-related neural responses. Wilroth \textit{et al.} \cite{wilroth2025eye} inferred attended-speech labels from eye-tracking features to supervise portable EEG-based speech-reconstruction models when ground-truth labels were unavailable. Kosmyna \textit{et al.} \cite{kosmyna2022target} evaluated low-channel EEG/EOG glasses and reported target-speaker detection performance comparable to a research-grade EEG headset. Xu \textit{et al.} \cite{xu2026eyeblink} further demonstrated that EOG-derived eyeblink sequences contain attention-related information and improve short-window AAD when combined with EEG.

However, existing work remains constrained by indirect eye-movement supervision and simplified EOG modeling. Consequently, the complementary relationship between EEG and EOG signals remains insufficiently modeled. These limitations motivate a unified multimodal AAD framework that explicitly captures modality-specific information and effective cross-modal interactions.

\section{Method}
This section details the proposed RAMamba-Net, a reliability-aware and Mamba-based multimodal fusion network for multimodal auditory attention decoding. We represent the EEG and EOG signals as \(\mathbf{X}_{\mathrm{EEG}}\in\mathbb{R}^{K\times C_{\mathrm{EEG}}\times T}\) and \(\mathbf{X}_{\mathrm{EOG}}\in\mathbb{R}^{C_{\mathrm{EOG}}\times T}\), respectively, where $K$ is the number of EEG frequency bands, $C_{EEG}$ and $C_{EOG}$ are the number of channels, and $T$ is the number of time points.

\begin{figure*}[htbp]
    \centering
    \includegraphics[width=1.0\linewidth]{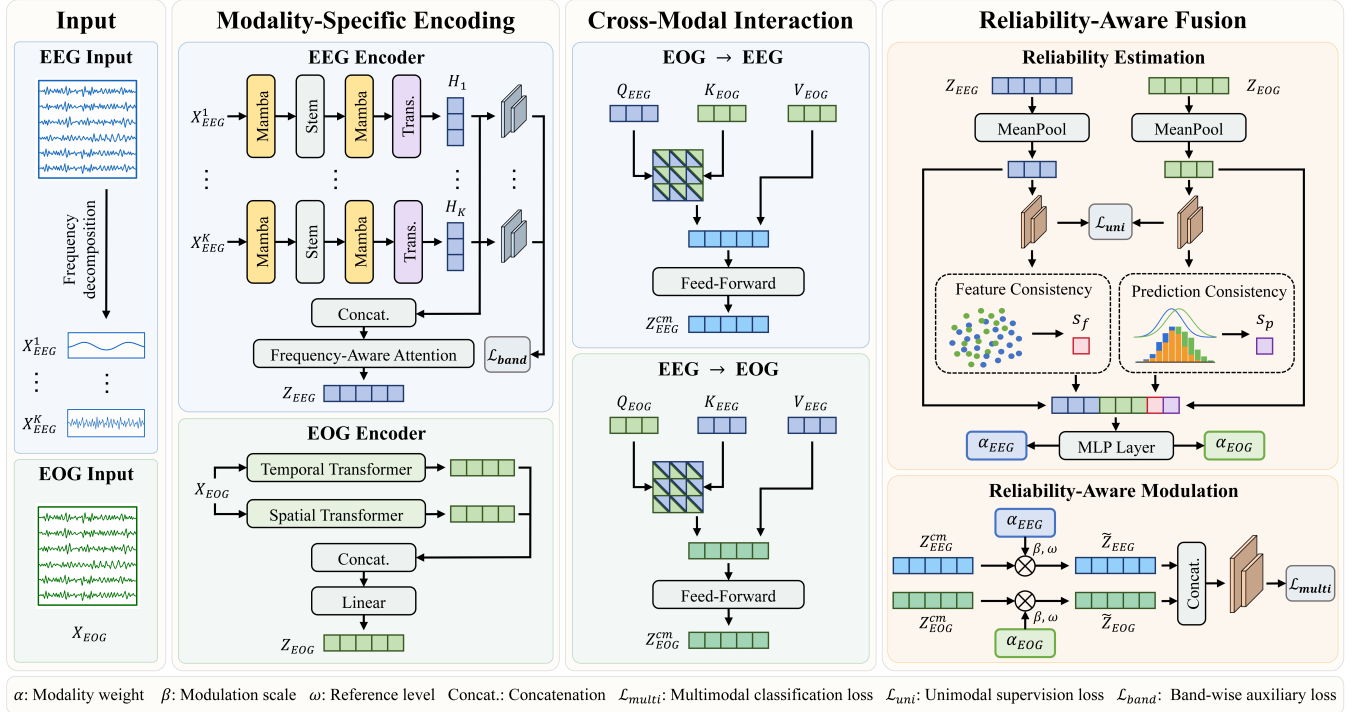}
    \caption{Overview of the proposed RAMamba-Net framework.}
    \label{fig:framework}
\end{figure*}

\subsection{Model Structure}
As illustrated in Figure~\ref{fig:framework}, RAMamba-Net comprises a backbone network and the RAM optimization module. The backbone network contains four components: an EEG encoder, an EOG encoder, a cross-modal fusion module, and classifiers. The two encoders first transform EEG and EOG signals into modality-specific feature representations. The fusion module promotes bidirectional interaction between the two modalities. The final fused feature vectors from each modality are concatenated and fed into the multimodal, unimodal, and band classifiers to identify the attended speaker. This backbone provides the basis for the reliability-aware learning strategy.

\subsubsection{EEG Encoder.}
The EEG encoder takes reference from the band-aware convolution Transformer architecture of FAConformer  \cite{wang2026faconformer}. We further enhance it by introducing the Mamba blocks, thereby constructing a Mamba-enhanced FAConformer. For the $k$-th EEG frequency band, the band-specific representation is obtained by:
\begin{equation}
\mathbf{H}_{k}=\mathcal{T}_{\mathrm{band}}^{k} \left(\mathcal{M}_{\mathrm{mid}}^{k} \left(\mathcal{S}^{k} \left(\mathcal{M}_{\mathrm{front}}^{k} \left(\mathbf{X}_{\mathrm{EEG}}^{k}\right)\right)\right)\right),
\end{equation}
where $\mathcal{M}_{\mathrm{front}}^{k}(\cdot)$ and $\mathcal{M}_{\mathrm{mid}}^{k}(\cdot)$ denote the front and middle Mamba blocks, respectively. $\mathcal{S}^{k}(\cdot)$ represents the band-specific stem module, and $\mathcal{T}_{\mathrm{band}}^{k}(\cdot)$ is the Transformer encoder. 

Given an input sequence $\{\mathbf{x}_{t}\}_{t=1}^{T}$, the Mamba blocks selectively propagate temporal information through:
\begin{equation}
\mathbf{h}_{t} = \bar{\mathbf{A}}_{t}\mathbf{h}_{t-1} + \bar{\mathbf{B}}_{t}\mathbf{x}_{t}, \ \ 
\mathbf{y}_{t} = \mathbf{C}_{t}\mathbf{h}_{t}.
\end{equation}
where $\mathbf{h}_{t}$ is the hidden state, $\bar{\mathbf{A}}_{t}$ denotes the discretized state-transition matrix, $\bar{\mathbf{B}}_{t}$ denotes the discretized input projection matrix, and $\mathbf{C}_{t}$ denotes the output projection matrix.
The front Mamba block $\mathcal{M}_{\mathrm{front}}^{k}(\cdot)$ reinforces the temporal structure of the raw band-specific EEG sequence, facilitating stable local spectro-temporal feature extraction by the stem, while the middle block $\mathcal{M}_{\mathrm{mid}}^{k}(\cdot)$ further enhances high-level feature dynamics to provide the Transformer with context-enriched token representations.

We concatenate the band-specific representations
\(\{\mathbf{H}_k\}_{k=1}^{K}\) along the band dimension and remap the result into a unified EEG token space:
\begin{equation}
\mathbf{Z}_{\mathrm{EEG}}
=\mathcal{R}_{\mathrm{EEG}}\left([\mathbf{H}_1 \,\Vert\, \cdots \,\Vert\, \mathbf{H}_K]\right),
\end{equation}
where \(\mathcal{R}_{\mathrm{EEG}}(\cdot)\) donates the frequency-aware attention module to adaptively determine the contribution of each frequency band to the final decision, promoting the cross-band interaction and dimensionality remapping.

\subsubsection{EOG Encoder.}
The EOG encoder takes reference from DBConformer \cite{wang2025dbconformer} and comprises parallel temporal and spatial branches. The temporal branch captures fine-grained temporal  dependencies through temporal Transformer. The spatial branch models inter-channel dependencies through spatial Transformer. Their outputs are formulated as:
\begin{equation}
\mathbf{Z}_{\mathrm{EOG}}^{\mathrm{temporal}} = \mathcal{E}_{t}\left(\mathbf{X}_{\mathrm{EOG}}\right),\ \ 
\mathbf{Z}_{\mathrm{EOG}}^{\mathrm{spatial}} = \mathcal{E}_{s}\left(\mathbf{X}_{\mathrm{EOG}}\right),
\end{equation}
where $\mathcal{E}_{t}(\cdot)$ and $\mathcal{E}_{s}(\cdot)$ denote the temporal and spatial branches.
The temporal and spatial representations are concatenated along the feature dimension and projected into the shared embedding space:
\begin{equation}
\mathbf{Z}_{\mathrm{EOG}}=\left[\mathbf{Z}_{\mathrm{EOG}}^{\mathrm{temporal}}\,\Vert\, \mathbf{Z}_{\mathrm{EOG}}^{\mathrm{spatial}}\right]\mathbf{W}_{r}+\mathbf{b}_{r},
\end{equation}
where \(\mathbf{W}_{r}\) and \(\mathbf{b}_{r}\) are learnable projection parameters. This dual-branch design preserves the temporal and spatial inductive biases of EOG while producing dimension-aligned tokens for subsequent cross-modal interaction.

\subsubsection{Cross-Modal Interaction.}
Given the modality-specific representations, we introduce cross-modal attention to establish explicit token-wise interactions between EEG and EOG. For each modality, its tokens serve as queries, while the other modality provides keys and values.

For modality \(m\in\{\mathrm{EEG},\mathrm{EOG}\}\), let \(\bar{m}\) denote the paired modality. The latent representation of modality \(m\) is enhanced by retrieving complementary information from modality \(\bar{m}\):
\begin{equation}
\mathbf{H}_{m\leftarrow\bar{m}}
=
\mathrm{softmax}\left(
\frac{\mathbf{Q}_{m}\mathbf{K}_{\bar{m}}^{\top}}{\sqrt{d_h}}
\right)\mathbf{V}_{\bar{m}},
\end{equation}
where \(d_h\) denotes the dimensionality of each attention head. The query, key, and value projections are defined as
\begin{equation}
\mathbf{Q}_{m}=\mathbf{Z}_{m}\mathbf{W}_{Q}^{m}, \ \ 
\mathbf{K}_{\bar{m}}=\mathbf{Z}_{\bar{m}}\mathbf{W}_{K}^{\bar{m}}, \ \ 
\mathbf{V}_{\bar{m}}=\mathbf{Z}_{\bar{m}}\mathbf{W}_{V}^{\bar{m}},
\end{equation}
where \(\mathbf{W}_{Q}^{m}\), \(\mathbf{W}_{K}^{\bar{m}}\), and
\(\mathbf{W}_{V}^{\bar{m}}\) are learnable projection matrices. The attention output is further refined through residual learning, layer normalization, and a feed-forward network:
\begin{equation}
\mathbf{Z}_{m}^{\mathrm{cm}} = \mathbf{Z}_{m} + \mathbf{H}_{m\leftarrow\bar{m}} + \mathrm{FFN} \left(\mathrm{LN} \left(\mathbf{Z}_{m} + \mathbf{H}_{m\leftarrow\bar{m}}\right)\right).
\end{equation}
The same update is applied in both directions. Each stream retains its modality-specific information while incorporating complementary evidence from the other modality. This symmetric interaction promotes balanced information exchange and more effective cross-modal alignment.

\subsection{RAM Optimization}
The backbone enables bidirectional EEG-EOG interaction but does not explicitly account for sample-wise variations in modality quality. RAM therefore infers the relative reliability of EEG and EOG from modality-specific representations that preserve their distinct signal characteristics, and adaptively regulates their contributions prior to classification. For each modality, the token sequence is mean-pooled into a sample-level descriptor for reliability estimation:
\begin{equation}
\mathbf{z}_{m}
=\mathrm{MeanPool}\big(\mathrm{stopgrad}(\mathbf{Z}_{m})\big).
\end{equation}
Mean pooling summarizes the modality-specific token sequence into a fixed-dimensional representation, while stop-gradient decouples reliability estimation from encoder optimization, preventing the RAM objective from directly reshaping the modality representations.

\subsubsection{Feature Consistency.}
Feature consistency measures the semantic agreement between EEG and EOG representations. It is computed from the pooled modality vectors using cosine similarity and normalized to $[0,1]$:
\begin{equation}
s_{f} = \frac{1}{2} \left( 1+\frac{\mathbf{z}_{EEG}^{\top}\mathbf{z}_{EOG}} {\|\mathbf{z}_{EEG}\|\,\|\mathbf{z}_{EOG}\|}\right).
\end{equation}
A larger $s_f$ indicates stronger feature-level agreement between the two modalities.

\subsubsection{Prediction Consistency.}
The pooled representations are further mapped to modality-specific predictions through lightweight unimodal classifiers $\mathcal{F}_{m}^{uni}(\cdot)$:
\begin{equation}
\mathbf{P}_{m} = \mathrm{softmax} \left(\mathcal{F}_{m}^{uni}(\mathbf{z}_{m})\right).
\end{equation}
Prediction consistency is derived from Jensen-Shannon divergence:
\begin{equation}
s_{p} = 1-\frac{D_{\mathrm{KL}}\left(\mathbf{P}_{EEG}\|\mathbf{M}\right)+D_{\mathrm{KL}}\left(\mathbf{P}_{EOG}\|\mathbf{M}\right)}{2\ln 2},
\end{equation}
where $\mathbf{M}=(\mathbf{P}_{EEG}+\mathbf{P}_{EOG})/2$, and $D_{\mathrm{KL}}(\cdot\|\cdot)$ denotes the Kullback-Leibler divergence. A larger $s_p$ indicates stronger decision-level agreement.

\subsubsection{Reliability-Aware Fusion.}
RAM combines the pooled modality-specific descriptors with feature-level and prediction-level consistency scores to construct a sample-wise reliability representation:
\begin{equation}
\mathbf{r}
=\left[\mathbf{z}_{\mathrm{EEG}}\,\Vert\,\mathbf{z}_{\mathrm{EOG}}\,\Vert\,s_f\,\Vert\,s_p\right],
\end{equation}
where \(s_f\) and \(s_p\) denote the feature and prediction consistency scores, respectively. The resulting representation jointly captures modality-specific information and cross-modal agreement for subsequent reliability estimation. Sample-wise modality weights are then estimated by:
\begin{equation}
[\alpha_{EEG},\alpha_{EOG}] = \mathrm{softmax} \left(\mathcal{G}_{\alpha}(\mathbf{r})\right),
\end{equation}
where $\mathcal{G}_{\alpha}(\cdot)$ is a reliability mapping MLP layer. The resulting weights regulate the cross-modally interacted representations:
\begin{equation}
\tilde{\mathbf{Z}}_{m} = \mathbf{Z}_{m}^{cm} \cdot \left[1+\beta(\alpha_{m}-\omega) \right],
\end{equation}
where $\beta$ controls the modulation scale, and $\omega$ denotes the reference level. The modulated EEG and EOG representations are directly concatenated and classified:
\begin{equation}
\hat{\mathbf{y}} = \mathcal{F}_{c}\left[\mathrm{vec}(\tilde{\mathbf{Z}}_{EEG})\,\Vert\,\mathrm{vec}(\tilde{\mathbf{Z}}_{EOG})\right],
\end{equation}
where $\mathrm{vec}(\cdot)$ denotes vectorization, and $\mathcal{F}{c}(\cdot)$ is the norm-constrained linear classifier.

\subsection{Training Objectives}
RAMamba-Net is optimized with a composite objective consisting of a multimodal classification loss, a unimodal supervision loss, and a band-wise auxiliary loss.
The multimodal classification loss supervises the final prediction obtained after cross-modal interaction and reliability-aware fusion:
\begin{equation}
\mathcal{L}_{multi}=\mathrm{CE}(\hat{\mathbf{y}},\mathbf{y}),
\end{equation}
where $\mathbf{y}$ denotes the ground-truth label, and $\mathrm{CE}(\cdot,\cdot)$ denotes the cross-entropy loss.
To preserve the discriminative capability of each modality-specific encoder, the unimodal loss directly supervises the auxiliary predictions from the EEG and EOG branches:
\begin{equation}
\mathcal{L}_{uni} = \mathrm{CE}(\mathbf{P}_{EEG},y) + \mathrm{CE}(\mathbf{P}_{EOG},y).
\end{equation}
The band-wise auxiliary loss further supervises each EEG frequency band through the independent lightweight classifier:
\begin{equation}
\mathcal{L}_{band} = \frac{1}{K} \sum_{k=1}^{K} \mathrm{CE}\Big( \mathcal{F}_k\big(\mathrm{vec}(\mathbf{H}_k)\big),y \Big),
\end{equation}
where $\mathcal{F}_k(\cdot)$ denotes the classifier for the $k$-th band.
The overall training objective is formulated as:
\begin{equation}
\mathcal{L} = \mathcal{L}_{multi} + \lambda_{uni}\mathcal{L}_{uni} + \mathcal{L}_{band},
\end{equation}
where $\lambda_{uni}$ controls the contribution of unimodal supervision. More specific details about the model structure of RAMamba-Net are provided in the supplementary material.

\section{Experiments and Results}
\subsection{Datasets}
We evaluate the proposed RAMamba-Net on two commonly used public AAD datasets.
\subsubsection{AVGC Dataset.}
AVGC is a public audio-visual dataset for AAD, comprising multimodal physiological and behavioral recordings \cite{rotaru2024what}. Subjects were required to attend to one of two competing speech streams, with the attended speaker located at $\pm 90^{\circ}$. The signals contain 64 EEG channels and 4 EOG channels at 8,192~Hz. Each trial lasted 10 minutes, with the attended direction switching after the first 5 minutes. We excluded subjects without public release permission and subjects with incomplete channel recordings. The final evaluation was conducted on 12 subjects.

\subsubsection{DTU Dataset.}
DTU is a public AAD dataset collected in simulated acoustic environments with different reverberation levels \cite{fuglsang2017noise}. It contains recordings from 18 normal-hearing participants listening to two competing speech streams, with two speakers presented from $\pm 60^{\circ}$ spatial directions. Each participant completed 60 dual-speaker trials, with each trial lasting approximately 50 seconds. The signals were recorded at 512~Hz and include 64-channel EEG and 6-channel periocular EOG. All 18 participants were included in the evaluation.

\subsubsection{Preprocessing.}
For AVGC, the recordings were downsampled to 128 Hz and band-pass filtered within 1-40 Hz. For DTU, EEG signals were high-pass filtered at 0.1 Hz, notch filtered at 50 Hz, and then resampled to 64 Hz. EEG additionally underwent drift removal, artifact attenuation, and average referencing, while the 6 EOG channels were directly aligned with the corresponding EEG samples after temporal preprocessing.

\subsection{Experimental Settings}
To keep each continuous recording segment within an independent subset, each subject's signals were chronologically partitioned before windowing. The first 90\% formed the training pool, while the remaining 10\% were reserved for testing. A validation set equal in size to the test set was then sampled from the training pool, and the remaining data were used for training. After data partition, each subset was independently segmented into 2-second decision windows. This resulted in an approximately 8:1:1 split across training, validation, and testing sets.

Common spatial pattern \cite{ang2008csp} was applied separately to EEG and EOG data to extract modality-specific spatial patterns. The filters were estimated from the training pool and applied to the validation and test sets. Given the Nyquist frequency after downsampling, AVGC and DTU were divided into 7 and 8 frequency bands, respectively. To comprehensively evaluate decoding performance, we report classification accuracy, balanced accuracy (BCA), Macro-F1, and Cohen's Kappa, covering overall discrimination, class-wise balance, and corrected agreement.

All models were implemented in PyTorch and trained on a single NVIDIA GeForce RTX 3090 GPU. Each experiment was repeated with the seed list $\{41,42,43,44,45\}$. We used Adam with an initial learning rate of $5\times10^{-4}$ and a weight decay of $3\times10^{-4}$. The batch size was set to 8 for AVGC and 16 for DTU. Each model was trained for up to 200 epochs, using early stopping with a patience of 10. The trade-off parameter $\mathcal{L}_{uni}$ was set to 0.8 for both datasets. The reliability module coefficients $\beta$ and $\omega$ were set to 0.1 and 0.5, respectively, and the number of Transformer layers and attention heads $H_c$ was set to 2.

\subsection{Unimodal Decoding}
Before studying EEG-EOG fusion, we first evaluated each modality separately to explore their unimodal discriminative ability. We compared three categories of representative deep AAD models. CNN-based models include EEGNet \cite{lawhern2018eegnet}, SCNN \cite{schirrmeister2017scnn}, and IFNet \cite{wang2023ifnet}. CNN-Transformer models include CTNet \cite{zhao2024ctnet}, TMSA-Net \cite{zhao2025tmsanet}, EEGConformer \cite{song2022conformer}, MSCFormer \cite{zhao2025mscformer}, MSVTNet \cite{liu2024msvtnet}, and DBConformer \cite{wang2025dbconformer}. AAD-specific models include DBPNet \cite{ni2024dbpnet}, DARNet \cite{yan2024darnet}, DHGCN \cite{zhou2025dhgcn}, and FAConformer \cite{wang2026faconformer}. In particular, DBPNet is not included in the EOG-only comparison, as it is specifically designed for 64-channel EEG with frequency band branches.

As reported in Table~\ref{tab:single_modality}, EEG consistently outperformed EOG across datasets and models, confirming its dominant role in AAD, while EOG remained clearly above chance level, indicating complementary attention-related information. FAConformer and DBConformer achieved the best EEG and EOG decoding performance, respectively, suggesting that frequency-aware temporal-spectral modeling better suits EEG, whereas temporal-spatial dual-branch modeling is more effective for EOG.

Overall, EEG provides stronger neural evidence, while EOG captures complementary ocular dynamics. Their distinct characteristics motivate modality-specific rather than shared encoders.

\begin{table}[htpb]
\centering
\setlength{\tabcolsep}{2.2pt}
\small
\begin{tabular}{c|cc|cc}
\toprule
\multirow{2.5}{*}{Model} & \multicolumn{2}{c|}{AVGC} & \multicolumn{2}{c}{DTU} \\
\cmidrule{2-5}
& EOG & EEG & EOG & EEG \\
\midrule
EEGNet & 58.67$_{\pm1.57}$ & 60.34$_{\pm0.91}$ & 60.81$_{\pm0.48}$ & 74.90$_{\pm1.43}$ \\
SCNN & 62.58$_{\pm0.84}$ & 67.74$_{\pm0.62}$ & 63.38$_{\pm0.54}$ & 81.45$_{\pm0.39}$ \\
IFNet & 61.94$_{\pm0.53}$ & 69.41$_{\pm0.43}$ & 62.73$_{\pm0.30}$ & \underline{82.32}$_{\pm0.34}$ \\
CTNet & 58.05$_{\pm1.32}$ & 67.41$_{\pm0.30}$ & 60.42$_{\pm0.44}$ & 74.84$_{\pm1.04}$ \\
TMSA-Net & 64.36$_{\pm0.51}$ & 73.58$_{\pm0.97}$ & 65.04$_{\pm0.42}$ & 80.47$_{\pm0.39}$ \\
EEGConformer & 63.70$_{\pm0.44}$ & 69.84$_{\pm0.68}$ & 59.10$_{\pm0.76}$ & 66.24$_{\pm0.70}$ \\
MSCFormer & 60.59$_{\pm0.92}$ & 69.14$_{\pm2.42}$ & 57.66$_{\pm0.54}$ & 65.35$_{\pm1.51}$ \\
MSVTNet & 61.30$_{\pm1.04}$ & 72.11$_{\pm2.46}$ & 59.60$_{\pm0.37}$ & 72.01$_{\pm1.80}$ \\
DBConformer & \textbf{67.30}$_{\pm1.43}$ & 69.06$_{\pm0.64}$ & \textbf{65.98}$_{\pm0.23}$ & 79.81$_{\pm0.31}$ \\
DARNet & \underline{65.36}$_{\pm0.28}$ & 72.83$_{\pm0.67}$ & \underline{65.42}$_{\pm0.50}$ & 82.24$_{\pm0.24}$ \\
DBPNet & -- & \underline{73.99}$_{\pm0.94}$ & -- & 79.83$_{\pm0.30}$ \\
DHGCN & 55.20$_{\pm1.90}$ & 63.61$_{\pm0.80}$ & 56.34$_{\pm0.26}$ & 78.23$_{\pm0.99}$ \\
FAConformer & 58.71$_{\pm1.00}$ & \textbf{74.23}$_{\pm1.07}$ & 62.88$_{\pm0.42}$ & \textbf{87.03}$_{\pm0.28}$ \\
\bottomrule
\end{tabular}
\caption{Average unimodal classification accuracies (\%) of 13 deep models on AVGC and DTU datasets.}
\label{tab:single_modality}
\end{table}

\subsection{Main Results}
Table~\ref{tab:main_results} compares RAMamba-Net with four unimodal and multimodal baselines, using DBConformer and FAConformer as the EOG-only and EEG-only backbones, respectively. Concat. directly combines the two modality-specific features, while RAMamba-Net w/o RAM retains cross-modal attention but removes the reliability-aware module.

\begin{table*}[htbp]
\centering
\small
\setlength{\tabcolsep}{1.2mm}
\begin{tabular}{c|cccc|cccc}
\toprule
\multirow{2.5}{*}{Approach} & \multicolumn{4}{c|}{AVGC} & \multicolumn{4}{c}{DTU} \\
\cmidrule{2-9}
& Accuracy & BCA & Macro-F1 & Cohen's Kappa & Accuracy & BCA & Macro-F1 & Cohen's Kappa \\
\midrule
EOG-only 
& 67.30$_{\pm1.43}$ & 67.58$_{\pm1.23}$ & 66.60$_{\pm1.56}$ & 0.35$_{\pm0.03}$ 
& 65.98$_{\pm0.23}$ & 65.47$_{\pm0.32}$ & 64.06$_{\pm0.28}$ & 0.30$_{\pm0.00}$ \\
EEG-only 
& 74.23$_{\pm1.07}$ & 74.49$_{\pm1.07}$ & 77.07$_{\pm1.14}$ & 0.58$_{\pm0.02}$ 
& 87.03$_{\pm0.28}$ & 87.62$_{\pm0.48}$ & 87.82$_{\pm0.31}$ & 0.75$_{\pm0.01}$ \\
Concat. 
& 74.74$_{\pm1.08}$ & 75.04$_{\pm1.12}$ & 77.59$_{\pm1.15}$ & 0.59$_{\pm0.02}$ 
& 87.19$_{\pm0.42}$ & 87.91$_{\pm0.35}$ & 87.98$_{\pm0.41}$ & 0.76$_{\pm0.01}$ \\
RAMamba-Net w/o RAM 
& \underline{77.16}$_{\pm0.98}$ & \underline{77.37}$_{\pm0.91}$ & \underline{79.55}$_{\pm0.96}$ & \underline{0.61}$_{\pm0.02}$ 
& \underline{87.90}$_{\pm0.54}$ & \underline{88.43}$_{\pm0.29}$ & \underline{88.76}$_{\pm0.57}$ & \underline{0.78}$_{\pm0.01}$ \\
RAMamba-Net (Full) 
& \textbf{79.99}$_{\pm1.03}$ & \textbf{80.23}$_{\pm0.66}$ & \textbf{80.51}$_{\pm0.56}$ & \textbf{0.64}$_{\pm0.01}$ 
& \textbf{88.42}$_{\pm0.15}$ & \textbf{89.21}$_{\pm0.16}$ & \textbf{89.04}$_{\pm0.18}$ & \textbf{0.80}$_{\pm0.00}$ \\
\bottomrule
\end{tabular}
\caption{Average classification performance of different approaches on AVGC and DTU.}
\label{tab:main_results}
\end{table*}

The results showed that EOG provides useful attention-related information, and EEG serves as a strong unimodal reference. Direct concatenation yielded only limited gains, indicating that simple feature fusion cannot fully exploit cross-modal complementary information. In contrast, explicit cross-modal interaction produced more consistent improvements by facilitating information exchange between EEG and EOG.

RAMamba-Net achieved the best results across all metrics and both datasets, with larger gains on AVGC. These results showed that cross-modal interaction and reliability-aware modulation jointly improve the accuracy, balance, and robustness of AAD by adaptively regulating modality contributions. Paired $t$-tests with adjusted $p$-values are provided in the supplementary material to verify statistical significance.

\subsection{Feature Visualization}
\begin{figure}[htpb]
\centering
\subfigure[AVGC]{\includegraphics[width=\linewidth]{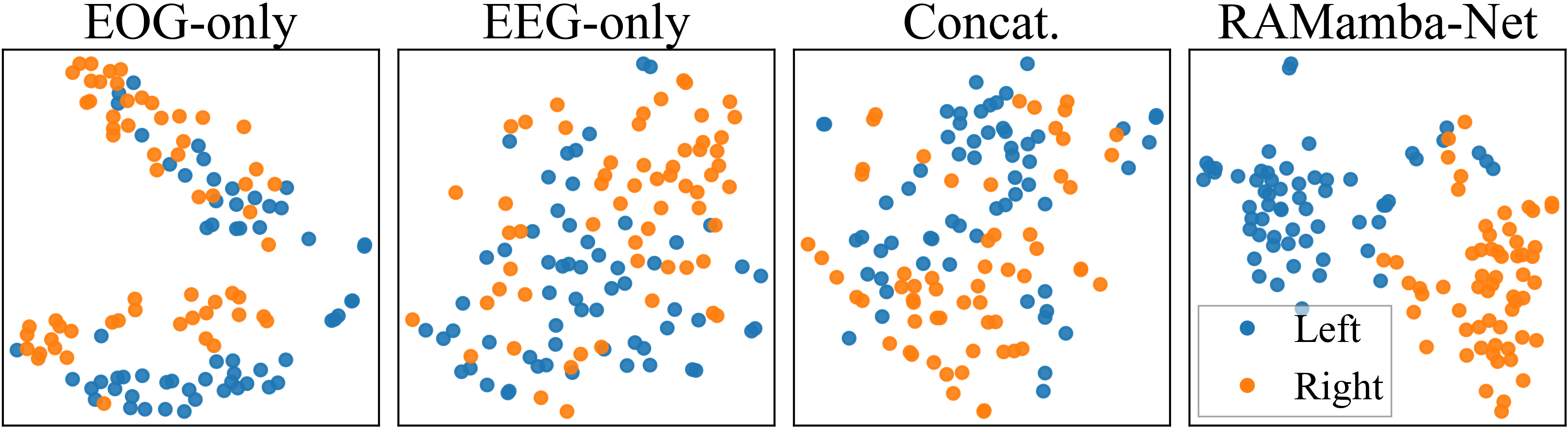}}
\subfigure[DTU]{\includegraphics[width=\linewidth]{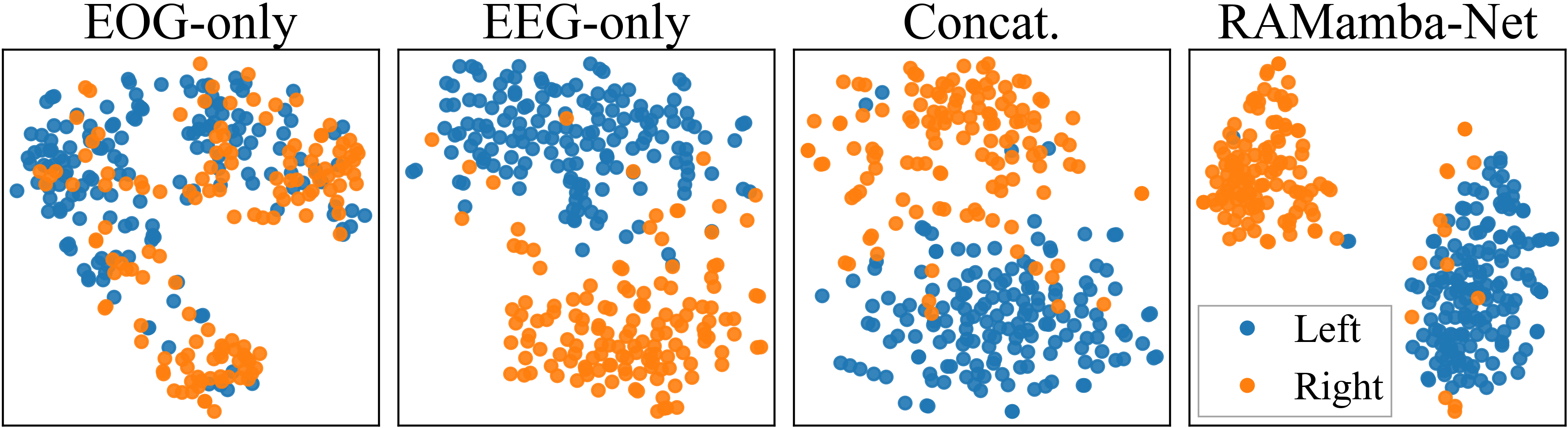}}
\caption{$t$-SNE visualization of feature distributions on AVGC and DTU.}
\label{fig:feature_visualization}
\end{figure}

To further examine how multimodal fusion affects representation learning, we visualized the features before classification using $t$-SNE \cite{van2008tsne}. We compared EEG-only, Concat., and RAMamba-Net to cover three representative stages: unimodal decoding, direct multimodal fusion, and reliability-aware multimodal fusion. This comparison can assess whether the proposed interaction and reliability mechanisms yield more discriminative feature distributions. The visualization results on AVGC and DTU are shown in Figure~\ref{fig:feature_visualization}.

The EEG-only features exhibited limited separability, with attention samples still mixed in the embedding space. Direct concatenation brings EOG information into the representation, but the feature distributions remained insufficiently organized, indicating weak cross-modal alignment.

RAMamba-Net yielded the clearest class structure on both datasets, with learned features forming more compact intra-class clusters and larger inter-class margins. The improvement was more evident on DTU, where the two attention classes are separated into well-structured regions. This suggests that RAMamba-Net produces more discriminative and robust feature representations than unimodal decoding and direct concatenation, effectively enhancing multimodal representation learning.

\subsection{Robustness to Noise Injection}
\begin{figure}[htbp]
    \centering
    \subfigure[AVGC]{\includegraphics[width=\linewidth]{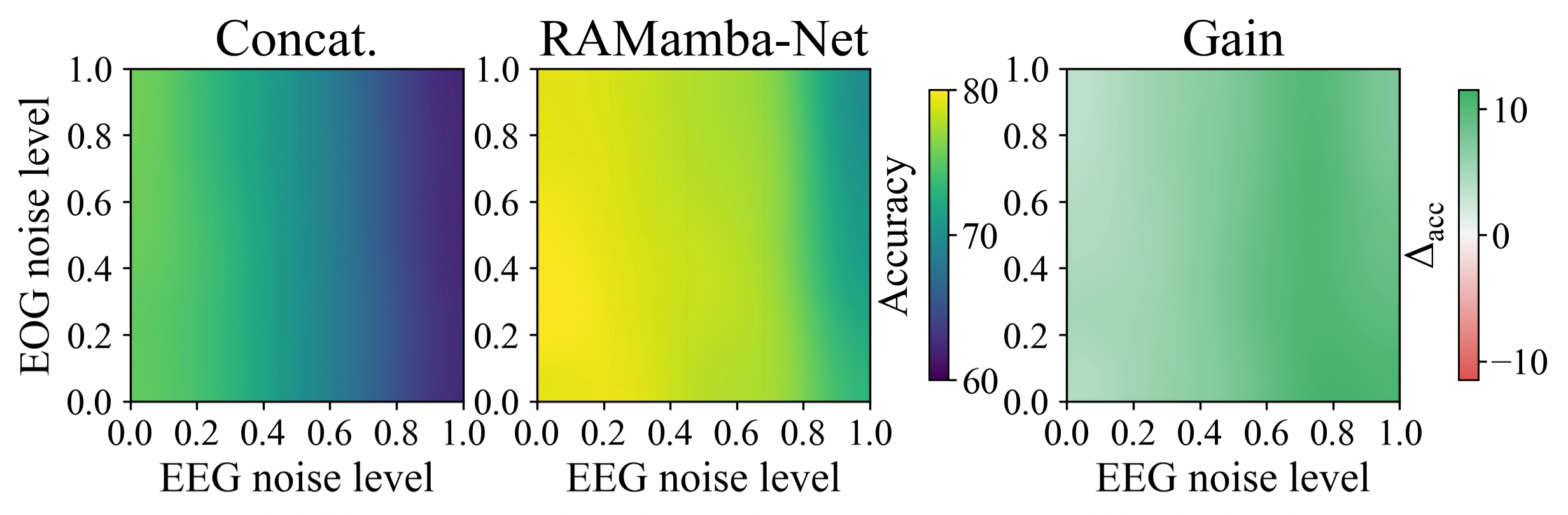}}
    \subfigure[DTU]{\includegraphics[width=\linewidth]{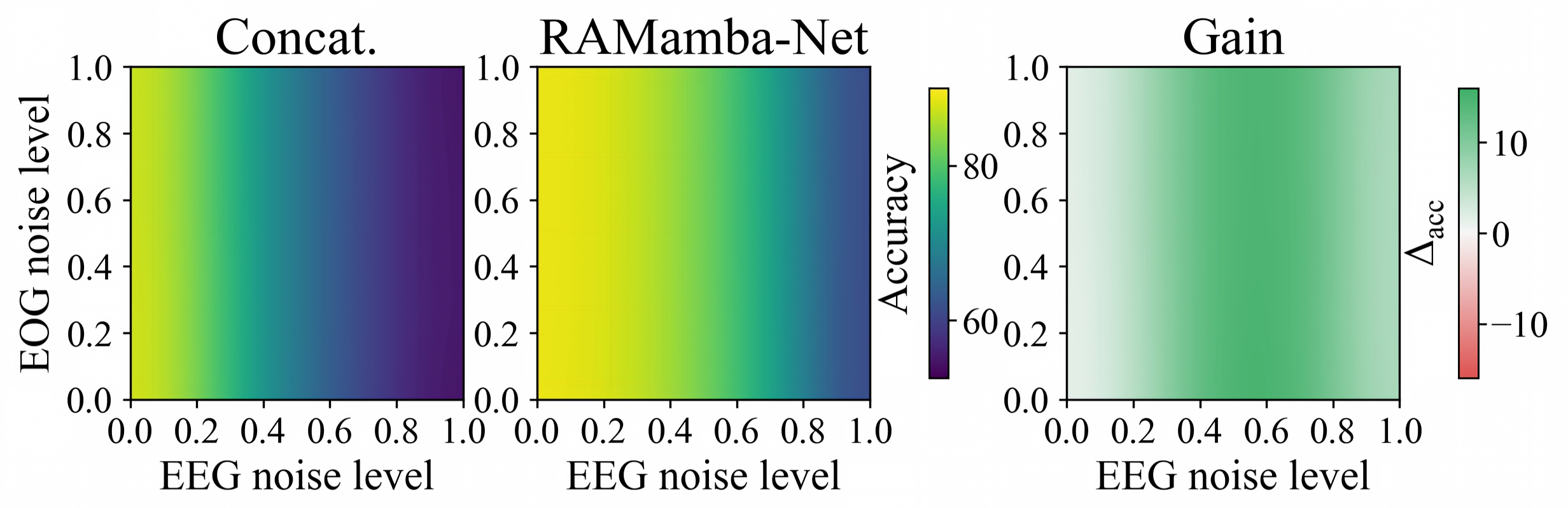}}
    \caption{Counterfactual robustness under modality-specific noise perturbations on AVGC and DTU.}
    \label{fig:noise_robustness}
\end{figure}

To evaluate robustness to modality degradation, we compared Concat. and RAMamba-Net under different EEG and EOG noise levels. The noise level of each modality was varied independently from 0 to 1. A value of 0 denotes a clean signal, while 1 indicates severe corruption with little clean information retained. The first two heatmaps reported classification accuracy under each noise combination. The last heatmap showed the accuracy gain. The results on AVGC and DTU are presented in Figure~\ref{fig:noise_robustness}.

With direct feature concatenation, accuracy declined mainly as EEG noise increased. The influence of EOG noise was less pronounced on both datasets. This suggests that the fusion process remains strongly dominated by EEG and is less robust when the primary modality is degraded. In contrast, RAMamba-Net maintained higher accuracy across nearly all noise conditions. The gain maps remained largely positive, even when both modalities were corrupted. This demonstrates that the proposed framework improves not only clean-signal performance but also robustness to changing signal quality. Bidirectional interaction preserves complementary cross-modal information, while RAM suppresses unreliable inputs and adaptively rebalances contributions from each modality. Together, these mechanisms enable more stable decoding under both modality-specific and joint noise perturbations.

\subsection{Ablation and Sensitivity Analysis}
\subsubsection{Ablation Study.}
To examine the contribution of each reliability component in RAM, we conducted an ablation study on modality preservation, feature consistency, and prediction consistency. Table~\ref{tab:component_ablation} reports the classification accuracy of different component combinations on AVGC and DTU.

\begin{table}[tb]
\centering
\small
\setlength{\tabcolsep}{1.4mm}
\begin{tabular}{c|ccc|c}
\toprule
\multirow{2}{*}{Dataset} & Modality & Feature & Prediction & \multirow{2}{*}{Accuracy} \\
& Preservation & Consistency & Consistency & \\
\midrule
\multirow{8}{*}{AVGC}
& $\times$ & $\times$ & $\times$ & 77.16$_{\pm0.98}$ \\
& $\checkmark$ & $\times$ & $\times$ & 78.40$_{\pm0.82}$ \\
& $\times$ & $\checkmark$ & $\times$ & 78.54$_{\pm0.64}$ \\
& $\times$ & $\times$ & $\checkmark$ & 77.89$_{\pm0.96}$ \\
& $\checkmark$ & $\checkmark$ & $\times$ & 78.59$_{\pm1.33}$ \\
& $\times$ & $\checkmark$ & $\checkmark$ & 79.08$_{\pm0.72}$ \\
& $\checkmark$ & $\times$ & $\checkmark$ & \underline{79.63}$_{\pm0.45}$ \\
& $\checkmark$ & $\checkmark$ & $\checkmark$ & \textbf{79.99}$_{\pm1.03}$ \\
\midrule
\multirow{8}{*}{DTU}
& $\times$ & $\times$ & $\times$ & 87.90$_{\pm0.54}$ \\
& $\checkmark$ & $\times$ & $\times$ & 88.05$_{\pm0.45}$ \\
& $\times$ & $\checkmark$ & $\times$ & 87.85$_{\pm0.39}$ \\
& $\times$ & $\times$ & $\checkmark$ & 87.95$_{\pm0.43}$ \\
& $\checkmark$ & $\checkmark$ & $\times$ & \underline{88.30}$_{\pm0.27}$ \\
& $\times$ & $\checkmark$ & $\checkmark$ & 88.03$_{\pm0.41}$ \\
& $\checkmark$ & $\times$ & $\checkmark$ & 88.08$_{\pm0.23}$ \\
& $\checkmark$ & $\checkmark$ & $\checkmark$ & \textbf{88.42}$_{\pm0.15}$ \\
\bottomrule
\end{tabular}
\caption{Ablation study of the proposed RAM modules on AVGC and DTU.}
\label{tab:component_ablation}
\end{table}

Modality preservation improved the backbone on both datasets, while the gains from feature and prediction consistency varied across datasets. Feature consistency led to the largest term gain on AVGC, whereas modality preservation remained the most effective on DTU. Pairwise combinations provided further improvements, and the complete RAM module achieved the highest accuracy on both datasets. These results show that the three terms capture complementary information of modality reliability. By preserving modality-specific information and jointly measuring consistency in the latent and prediction spaces, RAM enables more reliable sample-wise fusion and stronger multimodal AAD.

\subsubsection{Sensitivity Analysis.}
\begin{figure}[tb]
    \centering
    \includegraphics[width=1.0\linewidth]{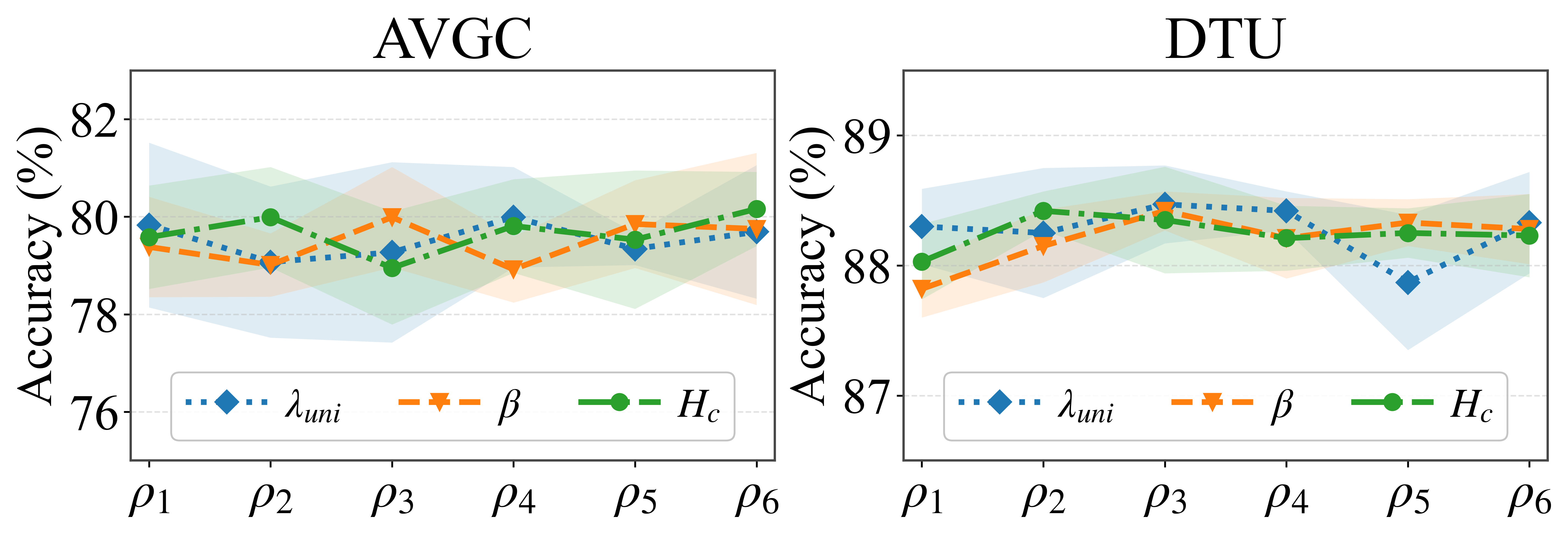}
    \caption{Sensitivity analysis on AVGC and DTU. The trade-off parameter $\lambda_{\mathrm{aux}}\in\{0.1,0.2,0.4,0.8,1.2,2.0\}$, the reliability module coefficient $\beta\in\{0.025,0.05,0.1,0.2,0.4,0.8\}$, and the number of attention heads $H_c\in\{1,2,4,8,16,32\}$.}
    \label{fig:parameter_sensitivity}
\end{figure}

We analysed the sensitivity of RAMamba-Net to three key hyperparameters: the trade-off parameter $\lambda_{\mathrm{uni}}$, the reliability module coefficient $\beta$, and the number of attention heads $H_c$ in the cross-modal Transformer encoder. Figure~\ref{fig:parameter_sensitivity} presents the parameter sensitivity analysis of RAMamba-Net on AVGC and DTU.

The accuracy curves remain stable across broad parameter ranges on both datasets. Varying $\lambda_{\mathrm{uni}}$ caused only moderate fluctuations, suggesting that auxiliary supervision does not require precise tuning. The model also maintained competitive performance across different values of $\beta$, showing limited sensitivity to the module strength. Similar stability was observed for $H_c$, as changing the number of attention heads produces no substantial performance degradation. These results demonstrate that RAMamba-Net is robust to hyperparameter variation and requires little dataset-specific tuning. The selected settings were adopted as representative settings within the stable performance ranges.

\section{Conclusion}
In this work, we proposed RAMamba-Net, a reliability-aware Mamba-based network for multimodal AAD using EEG and EOG. The model employs modality-specific encoders to preserve the distinct characteristics of the two signals. A Mamba-enhanced band-aware encoder captures continuous EEG dynamics, while a temporal-spatial dual-branch encoder models EOG patterns. Cross-modal attention enables explicit information exchange between the modalities. RAM further estimates sample-wise modality reliability, and uses the resulting weights to regulate fusion. Extensive experiments on AVGC and DTU demonstrate that RAMamba-Net consistently improves decoding performance, robustness, and representation quality through explicit cross-modal interaction and sample-wise reliability modeling. Future work will focus on improving the generalizability of EEG decoding under more challenging distribution shifts, including cross-subject and cross-dataset scenarios with diverse users and recording conditions. Moreover, efficient model adaptation and deployment strategies could be explored to reduce computational costs and enable practical multimodal AAD.

\clearpage
\appendix

\bibliography{aaai2027}

\end{document}